\documentclass[lettersize,journal]{IEEEtran}
\usepackage{amsmath,amsfonts}
\usepackage{algorithmic}
\usepackage{algorithm}
\usepackage{array}
\usepackage[caption=false,font=normalsize,labelfont=sf,textfont=sf]{subfig}
\usepackage{textcomp}
\usepackage{stfloats}
\usepackage{url}
\usepackage{verbatim}
\usepackage{graphicx}
\usepackage{cite}
\usepackage{booktabs}
\usepackage{multirow}
\usepackage{eso-pic}
\usepackage{xspace}

\usepackage{authblk}

\usepackage{hyperref}

\makeatletter
\def\eg{\emph{e.g.}\xspace}
\def\ie{\emph{i.e.}\xspace}

\makeatother

\newcommand{\omar}[1]{\textcolor{black}{#1}}
\newcommand{\ba}[1]{\textcolor{black}{#1}}

\begin{document}

\title{eMotion-GAN: A Motion-based GAN for Photorealistic and Facial Expression Preserving Frontal View Synthesis}




\author{Omar~Ikne,
        Benjamin~Allaert,
        Ioan~Marius~Bilasco,
        Hazem~Wannous,~\IEEEmembership{Senior Member,~IEEE}
\thanks{O. Ikne, B. Allaert and and H. Wannous are with
IMT Nord Europe, Institut Mines-Télécom, Univ. Lille, Centre for Digital Systems, F-59000 Lille, France. e-mail: \{omar.ikne,bejamin.allaert,hazem.wannous\}@imt-nord-europe.fr}
\thanks{I.-M. Bilasco is with Centre de Recherche en Informatique Signal et Automatique de Lille, Univ. Lille, CNRS, Centrale Lille, UMR 9189 - CRIStAL -, F-59000 Lille, France. e-mail: marius.bilasco@univ-lille.fr}
}

\markboth{}%
{Shell \MakeLowercase{\textit{et al.}}: A Sample Article Using IEEEtran.cls for IEEE Journals}

\maketitle

\begin{abstract}
Many existing facial expression recognition (FER) systems encounter substantial performance degradation when faced with variations in head pose.
Numerous frontalization methods have been proposed to enhance these systems' performance under such conditions. However, they often introduce undesirable deformations, rendering them less suitable for precise facial expression analysis.
In this paper, we present eMotion-GAN, a novel deep learning approach designed for frontal view synthesis while preserving facial expressions within the motion domain. 
Considering the motion induced by head variation as noise and the motion induced by facial expression as the relevant information, our model is trained to filter out the noisy motion in order to retain only the motion related to facial expression.
The filtered motion is then mapped onto a neutral frontal face to generate the corresponding expressive frontal face.
We conducted extensive evaluations using several widely recognized dynamic FER datasets, which encompass sequences exhibiting various degrees of head pose variations in both intensity and orientation. Our results demonstrate the effectiveness of our approach in significantly reducing the FER performance gap between frontal and non-frontal faces. Specifically, we achieved a FER improvement of up to +5\% for small pose variations and up to +20\% improvement for larger pose variations.
Code available at \url{https://github.com/o-ikne/eMotion-GAN.git}.
\end{abstract}

\begin{IEEEkeywords}
\omar{Facial expressions, frontal view synthesis, optical flow, motion warping, generative adversarial networks.}
\end{IEEEkeywords}

\section{Introduction}
\IEEEPARstart{F}{acial} Expression Recognition (FER) has received particular interest in recent years given its diverse real-life applications \cite{li2020deep}.
So far, effective methods have been proposed for FER in constrained environments and under ideal conditions, \eg, frontal pose and no illumination variation \cite{9039580}. However, when confronted with unconstrained environments, where individuals are free to move their heads, as it is the case in real-world applications, \eg, video surveillance, the FER task remains a challenge.

\begin{figure}[t]
    \centering
    \label{fig:intro_fig}
    \includegraphics[width=1.\linewidth]{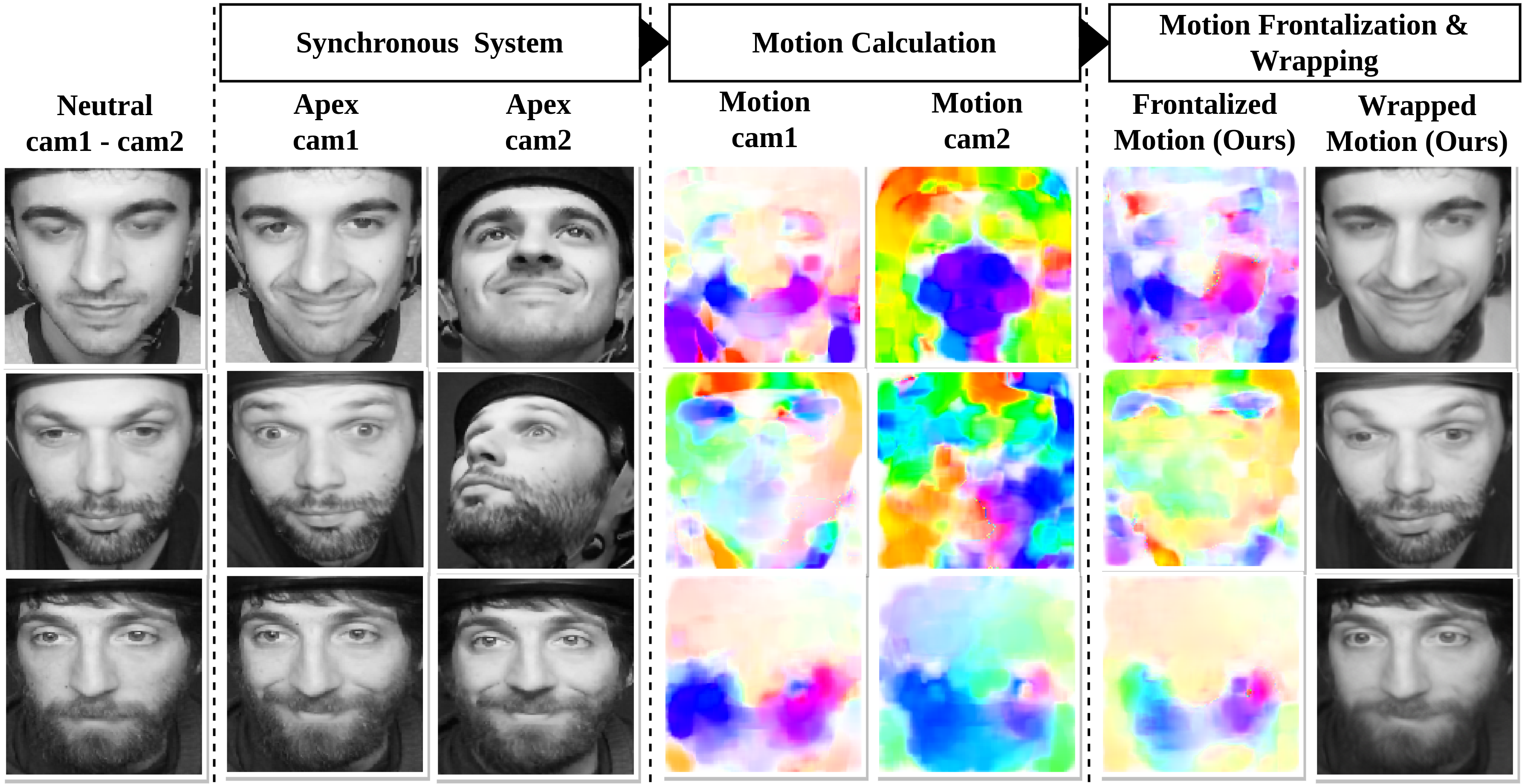}  
    \caption{eMotion-GAN estimates the motion induced by the variation of the head pose (column 5), to keep only the motion induced by the facial muscles. The latter is then transposed in the frontal plane (column 6) to match the ground truth (column 4) and warped into a neutral face (column 7) to assist in FER. The network training is based on a synchronous capture system (SNaP-2DFe \cite{allaert2018impact}), where facial motion is computed simultaneously in the absence (cam1) and presence (cam2) of head pose variation. 
    }
\end{figure}

FER can be performed in both motion and image (texture and geometry) domains. However, the image domain is more commonly used, as deep learning models can be trained on large-scale datasets. Nonetheless, FER models trained on images have shown a significant decrease in accuracy in the presence of head pose variations \cite{li2020deep}.
Although various Frontal View Synthesis (FVS) methods have been proposed to correct head pose variations and bring the face into a frontal view, they often introduce artifacts that can hamper the accuracy of facial expression recognition models \cite{allaert2018impact}, given that these methods are mainly used in the field of face recognition \cite{hu2018pose}, where the appearance of deformations causes minor performance issues. 
However, they are not well suited to facial expression recognition, where preservation of facial geometry is very important to correctly characterize facial movements.
Thus, exploring the motion domain is worthwhile. \ba{Recent research tends to exploit optical flow to ensure the proper quality of the frontalization process \cite{dong2020supervision}. Based on this observation, the integration of a constraint driven by the movement induced by the bio-mechanical constraints of the face, seems a promising solution for conditioning the frontalization process in order to preserve facial expressions.} We suggest considering head pose variation as noise in the motion domain and focus on extracting facial motion while filtering out head pose motion. This approach offers the advantage of preserving consistent motion patterns for the same facial expression, which remain similar across individuals, regardless of their age, gender, or ethnicity. \ba{The extracted motion is used to control the frontalization process. The reconstructed frontal face preserves the bio-mechanical properties, and guarantees the recovery of the initial facial expression (Figure \ref{fig:intro_fig}) for improved FER accuracy.} 

Given the rise of generative models and their ability in tackling denoising and style transfer challenges, we explore a solution based on conditioned Generative Adversarial Networks (cGANs) \cite{mirza2014conditional} to address this issue. Our novel approach brings forth a range of distinctive features, such as:

\textbf{\textit{Unconstrained inter-individual variation -}} This strategy takes advantage of the similarity of facial motion to reduce inter-individual variation for the same facial expression.

\textbf{\textit{Facial landmarks independence -}} In contrast to recent approaches based on face texture and geometry, the proposed approach do not rely on landmarks or other facial attributes and is therefore not strongly impacted by the accurate location of facial landmarks.

\textbf{\textit{Expression transfer ability -}}
By generating facial motion flows instead of face images, this approach gains in flexibility. The ability to warp facial motion to any face, real or synthetic, extends its utility to a variety of face categories, including people, animals and drawings, and can serve as a useful data augmentation technique.

\ba{The structure of this paper is as follows: in Section \ref{sec:related_work}, we review relevant recent methods for FER and the FVS methods for dealing with the variation in head pose. Section \ref{sec:approach} present our generative model to deal with the head pose variation while preserving facial expressions. In Section \ref{sec:exp}, the experimental setup including datasets, protocols and classification models are detailed. The experimental results are provided in Section \ref{sec:eval}. We conclude and discuss future work in Section \ref{sec:conclusion}.}

\section{Related Work}
\label{sec:related_work}

\subsection{Frontal View Synthesis}
Existing methods for FVS can be roughly categorized into two main groups: features-based and learning-based methods.

\textit{Features-based methods -} belong to a class of techniques primarily founded on geometric transformation principles that leverage texture and/or geometry. These methods have proven effective in dealing with pose variations in the frontal plane, offering solutions such as rigid alignment achieved through landmark-based techniques \cite{zarbakhsh20204d}, non-rigid alignment utilizing optical flow \cite{liu2008sift}, and even a combination of rigid transformation and non-rigid deformation for more refined transformations \cite{kang2023expression}, \cite{kang2021robust}. These techniques demonstrate considerable success when it comes to inducing transformations for in-plane variations. However, the challenge arises when handling out-of-plane (3D) pose variations, where the existing features-based methods fall short. To address this limitation, recent advancements have been proposed, such as the incorporation of facial landmarks and thin-plate splines warping \cite{al2020automatic}, or the utilization of the 3D Morphable Model (3DMM) approach \cite{zhu2015high} to estimate the pose of a non-frontal face relative to a 3D frontal face model, then use the pose coordinates to warp the pixels of the input face onto a frontal face. Despite these efforts, many of these methods require model adjustments and manual annotation of out-of-plane facial landmarks, making the accuracy of these landmarks a crucial factor for achieving effective face frontalization \cite{belmonte2021impact}. Moreover, the process of filling in invisible regions due to pose assumptions is based on the assumption of face symmetry, which, in reality, is not always valid due to variations in illumination and occlusions, raising further challenges to the accuracy of frontalization.

\textit{Learning-based methods -} represent another major group of FVS techniques that rely on generative models, most notably Variational Auto-Encoders (VAEs) \cite{kingma2013auto} and Generative Adversarial Networks (GANs) \cite{yin2017towards}.
While VAEs offer relatively easier learning processes, GANs stand out as more suitable for addressing the highly challenging task of frontal view synthesis, thanks to their enhanced flexibility in incorporating constraints during the reconstruction process. 
The 3DMM field has also witnessed numerous extensions mainly motivated by the capabilities of GANs, leading to a multitude of widely proposed methods \cite{yin2017towards},\cite{huang2017beyond},\cite{wu2020cascade}, \cite{zhao2020recognizing}, \cite{lee2022exp}. 
Recent breakthroughs in texture-based approaches have made it possible to correct pose variations while preserving facial expressions \cite{zhang2018face,ju2022complete}. 
Nevertheless, it's essential to acknowledge that texture-based methods require extensive datasets comprising a large number of subjects and samples to yield satisfactory results in face-frontalization tasks, as they rely essentially on image texture and geometry, both of which exhibit significant variation between individuals. 
Conversely, some researchers have proposed geometric transformation methods that rely on generative models (GANs) and optical flows to address the challenges of frontal view synthesis \cite{chang2021learning}, \cite{peng2023facial}.
\ba{Chang et al. \cite{chang2021learning} use optical flow to train a network to de-express towards a neutral face by suppressing facial motion and re-express to reconstruct the original face. 
This work highlights the benefits of optical flow for correcting head pose and transferring expression to the face.
However, the authors highlight that unsupervised learning tends to perform less well than existing supervised learning.
Other approaches \cite{koujan2020deepfaceflow,peng2023facial} exploit optical flow to enhance the quality of facial reconstruction in the frontal plane. 
Although these methods improve on the performance of current systems, they rely on precise landmarks localization and 3D face shape estimation to correctly frontalize the face. 
However, the combination of these different features tends to drastically reduce the performance of these methods in the presence of large head pose variations or facial expressions.}

It's worth mentioning that while the majority of learning-based approaches focus primarily on improving face recognition performance \cite{koujan2020deepfaceflow,peng2023facial}, they may not explicitly prioritize the preservation of facial expressions during the face frontalization process. This aspect requires particular attention in order to guarantee a holistic and natural synthesis of the face view that accurately represents the facial expressions.
\omar{Incorporating an FER-related loss as an integral component of the overall reconstruction loss can be an effective way of ensuring the preservation of facial expression dynamics during the frontalization process.}

\subsection{\omar{Training strategies and datasets}}

Training learning-based models, particularly generative models, can be realized through three main strategies. The first, supervised learning, necessitates an extensive dataset containing paired face images depicting non-frontal view faces in conjunction with their corresponding frontal view counterparts, commonly denoted as "ground truths" \cite{yin2017towards,huang2017beyond}.
The second approach involves semi-supervised learning, in which the dataset includes both annotated data with ground truth and unannotated data \cite{zhang2020semi}. The key aspect of these methods is the use of adversarial identity-preserving loss for efficient learning. The third method is unsupervised learning, which overcomes the need for paired data and relies primarily on cycle consistency \cite{zhu2017unpaired}. Zhou et al. \cite{zhou2020rotate} proposed to synthesize profile views from a collection of frontal views in order to generate input versus target pairs for training image-to-image translation GANs.

While semi-supervised and unsupervised learning aim to reduce the need for paired data, supervised learning remains a reliable method, particularly when paired data is available, and offers distinct advantages over semi-supervised and unsupervised learning approaches in image-to-image translation problems \cite{chang2021learning}. With access to paired data, supervised learning has well-defined performance metrics, making it easier to measure and optimize model performance. Which results in a greater predictive accuracy and a better generalization to unseen data when the training dataset is diverse and representative.

\begin{figure}[!b]
    \centering
    \label{fig:related_fig}
    \includegraphics[width=0.98\linewidth]{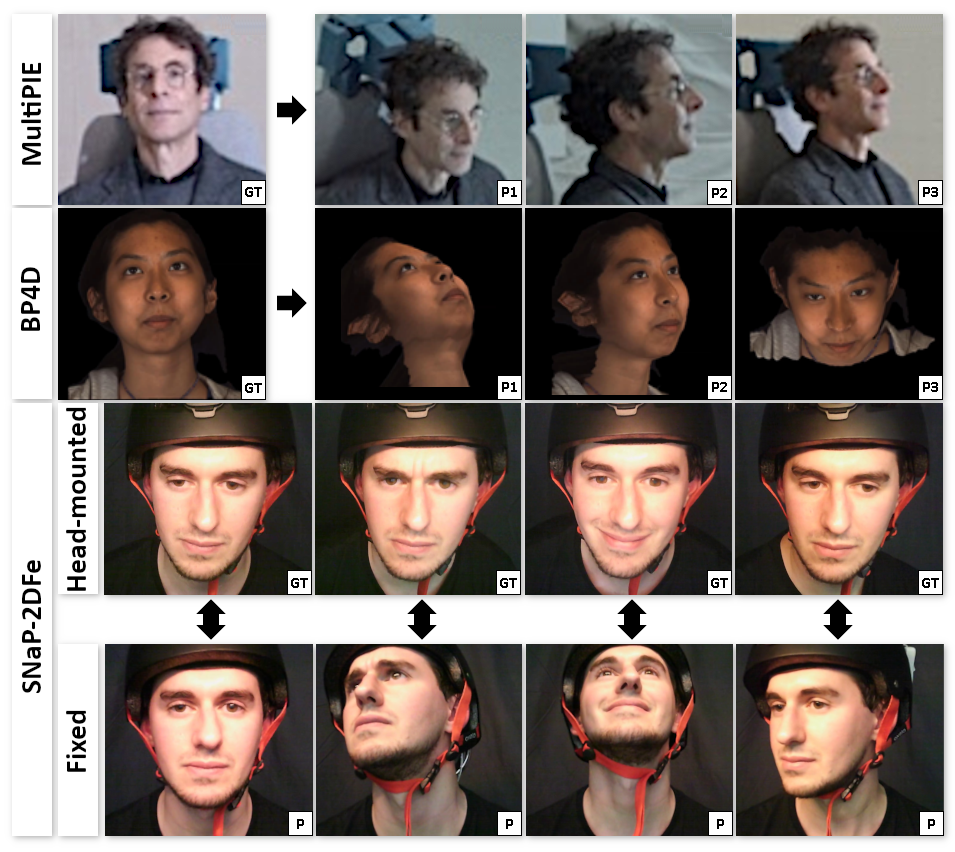}  
    \caption{\ba{Setup used in datasets to generate head pose variations and provide ground truth of facial movement and expression. GT: Ground Truth; P: Pose.}}
\end{figure}

\ba{The training of supervised approaches to face frontalization has been based mainly on the MultiPIE \cite{gross2010multi} and BP4D \cite{zhang2014bp4d} databases. However, the strategies employed to provide varied poses are not always well adapted, as illustrated in Figure \ref{fig:related_fig}. Multi-Pie relies on a multi-view system, which involves taking an image of a face from several angles. BP4D involves recording frontal faces using a 3D capture system. Variations in pose are generated automatically by rotating the camera around the 3D model. Although these datasets tend to provide a wide range of head pose variations, exploiting the frontal face as ground truth, the data acquired do not reflect real pose variations, where the head is constrained by bio-mechanical rules. These constraints guarantee consistency between head movement and facial expressions, which are not present in most of these datasets. Among the many datasets, SNaP-2DFe \cite{allaert2018impact} stands out for the synchronized capture system, offering simultaneous video acquisition of the expressive face with and without variations in head pose.} These datasets offer sequences of basic facial expressions played out, with variations in head pose that closely resemble real-world scenarios. This crucial feature ensures that the dataset accurately captures the complexities of facial expressions and bio-mechanical constraints in natural environments, making it particularly well-suited to meet the challenges of head pose variation in FER tasks. Consequently, the constraint of no paired data is minimal and can be overcome efficiently.



\begin{figure*}[t]
    \centering
    \includegraphics[width=0.9\textwidth]{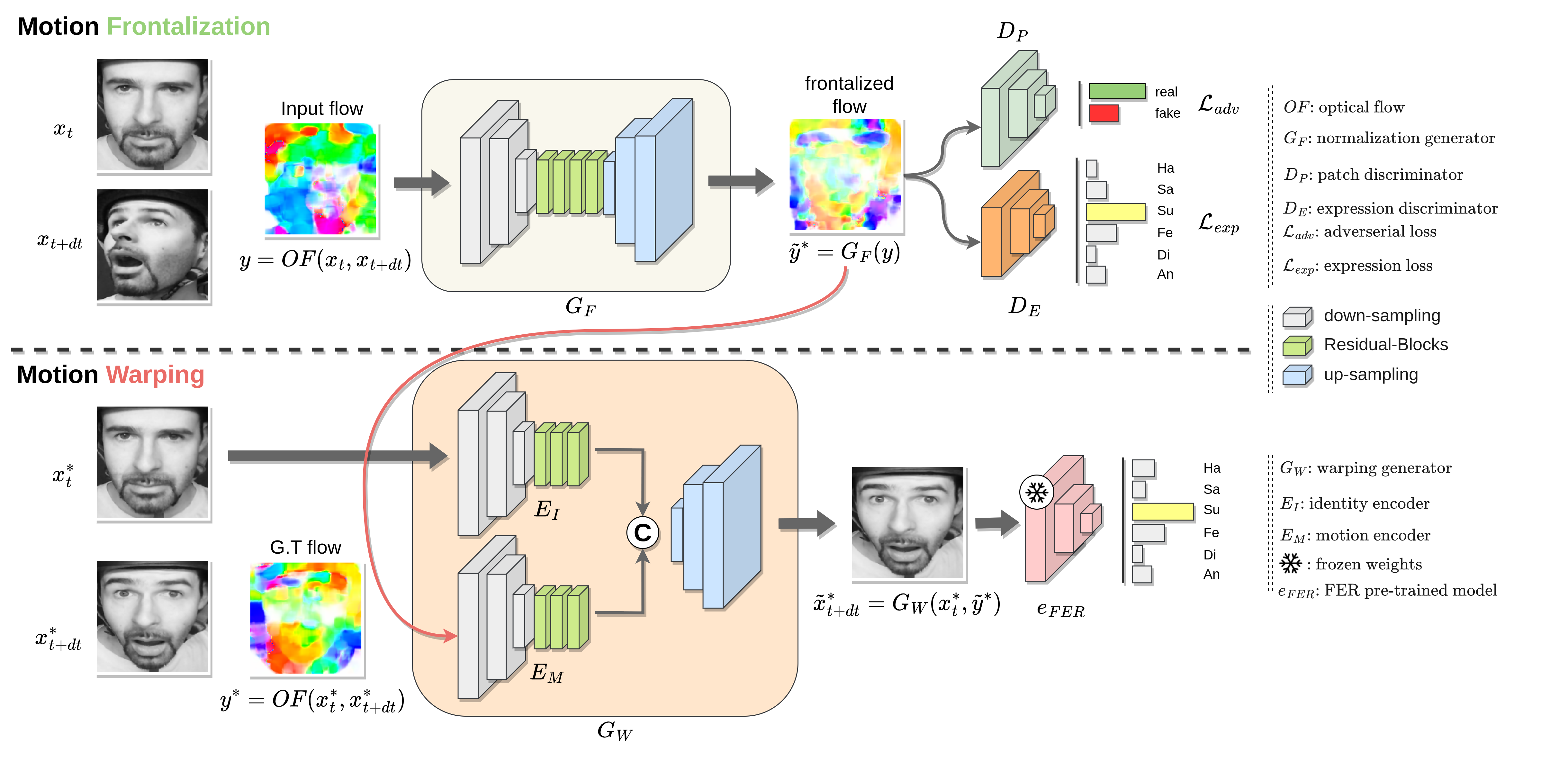}
    \caption{\textbf{eMotion-GAN approach}.
    $1^{st}$ phase: motion frontalization. Given an input optical flow ($y$), the generator $(G_{F})$ estimates and filters the motion induced by the variation of the head pose and transposes the motion induced by the facial expression in the frontal plane ($\tilde{y}^*$) to approximate the real motion ($y^*$). Besides the reconstruction losses, the expression discriminator $D_{E}$ is introduced to force $G_{F}$ to preserve the facial expression through the loss $\mathcal{L}_{e}$. The discriminator $D_{P}$ ensures that the properties of the facial movement (\eg intensity and direction) are preserved through the $\mathcal{L}_{adv}$ loss. $2^{nd}$ phase: motion warping. Given a neutral face $(x_t^*)$ and the frontalized motion field $\tilde{y}^*$, the generator $G_{W}$ generates the corresponding expressive face. A pre-trained classifier $e_{FER}$ predicts the corresponding expression.}
    \label{fig:approach_illustration}
\end{figure*}


\section{eMotion-GAN}
\label{sec:approach}
We propose an end-to-end deep learning approach for face FVS based in the motion domain. It consists of two main phases: a) motion frontalization and b) motion warping as illustrated in Figure \ref{fig:approach_illustration}. 

Let $X \subset \mathbb{R}^{W\times H\times C}$ be a source domain, \eg, non-frontal faces with facial expressions, and $X^* \subset \mathbb{R}^{W\times H\times C}$ the corresponding target domain, \eg, frontal faces with same facial expression, where $H$, $W$, and $C$ denote the image height, width, and channel number respectively.
We aim at learning a non-linear 2D-to-2D mapping $G: X \mapsto X^*$ that brings any arbitrary non-frontal face into a frontal view with the additional constraint of preserving facial expressions, \ie $\forall (x, x^*) \in X\times X^*, ~ e(G(x)) = e(x^*)$ where $e: X_f\mapsto \mathcal{E}$ is a mapping of frontal faces $X^*$ into the set of facial expressions $\mathcal{E}$. The mapping $G$ can be roughly written as $G = G_{W} \circ G_{F}$, where $G_{F}(\cdot)$ is a motion frontalization mapping and $G_{W}(\cdot)$ is a motion warping mapping.

Considering the recent promising results for image denoising \cite{chen2018image} and style transfer \cite{Xu_2021_ICCV} problems using GAN-based models, we propose to take advantage of this architecture to solve the problem at hand.


\subsection{Motion Calculation}
Given a pair of consecutive face frames $x_t,~ x_{t+dt} \in X$ with a facial expression, the optical flow which specifies the \emph{magnitude} and \emph{direction} of pixel motion between the two frames is calculated $y = OF(x_t, x_{t+dt}) \in Y$, where $Y\subset \mathbb{R}^{W\times H\times 2}$ is the set of the optical flows and motion patterns of the expressive faces. 


In order to facilitate the generalization of the model to handle various motion intensities, the magnitude of motion is normalized by a common \emph{clipping} technique. The clipping threshold can be chosen efficiently by plotting the distribution of the magnitude over the dataset used and then choosing a value that prevents large information loss, \eg, the third quartile $Q_3$. The calculation process is as follows: first the face is detected, cropped, then the optical flow is calculated, on high resolution images to avoid large information loss, and resized to match the model input.
A larger cropping less focused on the inside of the face allows a better observation of the head movements.

\subsection{Motion Frontalization}
Given an optical flow $y\in Y$ of non-frontal expressive face, $G_{F}$ estimates and filters the motion induced by the variation of the head pose and transposes the motion induced by the facial expression in the frontal plane $\tilde{y}^* = G_{F}(y)$ to match the ground truth flow $y^*\in Y^*$.

In the following, we present our proposed cGAN architecture for the frontalization model, which is mainly based on a flow generator $G_{F}$ and two discriminators, a motion discriminator $D_{P}$ and an expression discriminator $D_{E}$.

\paragraph{Flow Generator ($G_{F}$)} 
The underlying architecture compromise an encoder network and a decoder network with residual blocks as bottleneck. The input and the output of the network are set to  $128\times128\times2$.

The encoder consists of 4 downsampling blocks, each with a 2D convolution, an instance normalization \cite{ulyanov2017improved}, and a ReLU. 
Reflection padding is added in the first layer to prevent checkerboard artifacts \cite{niklaus2017video}. Instance normalization is applied to all features of a channel, allowing the training process to achieve lower loss levels than batch normalization \cite{ulyanov2017improved} over a wide range of small batch sizes. Two residual blocks \cite{he2016deep} are added to further process the encoded flow. The residual blocks involve a 2D convolution, an instance normalization and a ReLU.
The decoder processes the encoded motion with 5 residual blocks and reconstructs the frontalized motion with 4 upsampling, each with a 2D convolution, a layer normalization with affine transformation parameters and a ReLU.


\paragraph{PatchGAN Discriminator ($D_{P}$)} 
Inspired by the Pix2Pix GAN discriminator \cite{isola2017image}, we design our discriminator based on the PatchGAN approach. By operating at the patch level, the PatchGAN discriminator achieves higher computational efficiency, as it requires fewer parameters compared to those operating at the image level. In addition, image-level discriminators may suffer from overconfidence, \ie if they rely on a small set of features to distinguish true from fake images, the generator may exploit these features to fool the discriminator. Although control techniques such as dropout or label smoothing can help alleviate this problem, the PatchGAN discriminator is better suited since it discriminates at the patch level, thus avoiding this problem.

Each block of the proposed architecture consists of a 2D convolution layer with a kernel size of 4 and a stride size of 2, an instance normalization and a Leaky ReLU which makes the gradients less sparse.
    
\paragraph{Expression Discriminator ($D_{E}$)} 
We introduce a second discriminator to force the flow generator $G_{F}$ to comply with the additional constraint of preserving facial expressions while reconstructing the frontal flows. It serves as an optical flow CNN-based classifier (Flow-CNN) for FER. It takes an optical flow which encodes the facial motion of a subject, as input and predicts the corresponding facial expression.
Its architecture consists of 3 blocks for features extraction, each composed of a 2D convolution, a batch normalization, a max pooling and finally an MLP layer for classification.

\subsection{Motion Warping}
The motion warping phase consists in warping the frontalized facial motion synthesized by $G_{F}$ onto a neutral frontal face to generate the corresponding expressive frontal face.
In this context, given a neutral face image $x_t$ and a frontalized optical flow $\tilde{y}^*=G_{F}(y)$, the motion warping model aims at reconstructing the corresponding expressive face by applying the given motion field to the neutral face, \ie
$\tilde{x}^{*}_{t+dt} = G_{W}(\tilde{y}^*, x^*_t)$.
Where $G_W$ denotes the image generator. 
The expressive frontal faces produced are examined using an image-based facial expression discriminator, called $e{FER}$, which guarantees the fidelity of facial expression preservation in the image domain.

\paragraph{Image Generator ($G_{W}$)}
The image generator includes a motion encoder, an identity encoder for encoding the identity of the subject's face, and a joint decoder for applying the encoded motion to the encoded face identity.

The architectures of encoders and decoder are roughly similar to those of the flow generator with a difference of number of sampling and residual blocks. Both encoders compromise 3 downsampling and 3 residual blocks. While the decoder consists of 3 residual and 3 upsampling blocks.

\paragraph{Expression Discriminator ($e_{FER}$)}
For the FER task on images synthesized by the image generator, we use a pre-trained model, namely DMUE \cite{she2021dive}, with the design of latent distribution mining and pairwise uncertainty estimation. 
DMUE demonstrates excellent performance in FER due to its specific design to meet the challenge of annotation ambiguity \cite{she2021dive}. Moreover, it provides valuable feedback to the image generator, indicating whether the facial expression has been accurately warped into the face or not.

\subsection{Loss Functions}

For a standard conditional GAN, the objective function which refers to the conditional adversarial loss function is formulated as \cite{mirza2014conditional}:
\begin{equation}
\begin{split}
\mathcal{L}_{cGAN}(G, D) & = \mathbb{E}_{X, Y}[\log{D(Y)}] \\
& + \mathbb{E}_{X, Y}[\log{(1 - D(X, G(X, Z)))}] 
\end{split}
\end{equation}
where the model is trying to learn the mapping $G:\{X, Z\}\mapsto Y$, $X$ is the source domain, $Y$ is the target domain and $Z$ is the noise sampled from a normal distribution. The generator tries to minimize this loss while the discriminator tries to maximize it. In our case, the loss of the discriminator compromises the losses of both motion and expression discriminators.

Since GANs generally suffer from poor quality generation and artifacts, we propose to add additional penalties to the loss function to impose constraints on the quality of the generated flows and images. 

For the image generator, the loss function consists of three main parts: an MAE loss, a VGG-based perceptual loss \cite{Simonyan2014VeryDC} and an image-based facial expression loss.
The corresponding total loss is given by:
\begin{equation}
    \mathcal{L}_{G_{W}} = \alpha_1\mathcal{L}_{1} + \alpha_2\mathcal{L}_{perc} + \alpha_3\mathcal{L}_{e_{FER}}
\end{equation}
$\mathcal{L}_{e_{FER}}$ is the categorical cross-entropy loss between the true and predicted facial expressions. While $\alpha_1$, $\alpha_2$ and $\alpha_3$ are loss scaling factors. For the flow generator, the generated flows can be improved by using different loss functions to preserve the properties of the facial motion (movement pattern, magnitude and intensity), and to recognize the facial expressions. Two main losses, the End Point Error (EPE) and the Charbonnier loss

\paragraph{EndPoint Loss} 
Is a common optical flow metric that prioritizes accuracy and indirectly favors smoothness by penalizing large errors. It's calculated by measuring the Euclidean distance between an estimated flow, denoted as $G_{F}(y)$, and the ground-truth flow $\tilde{y}^*$.
\begin{equation}
    \mathcal{L}_{epe} = \mathbb{E}_{y, \tilde{y}^*}\left[\|\tilde{y}^*-G_{F}(y)\|\right]
\end{equation}

\paragraph{Charbonnier Loss} 
Is a differentiable variant of $\mathcal{L}_{1}$ that is useful for flow and depth estimation tasks that require robustness. Its particularity lies in the fact that it behaves as a quadratic loss near the origin and as an absolute loss far from the origin.
\begin{equation}
    \mathcal{L}_{char} = \mathbb{E}_{y, \tilde{y}^*}\left[\|\sqrt{\mathcal{L}_{1}(\tilde{y}^*, G_{F}(y))^2 + \epsilon^2}\|\right]
\end{equation}
$\epsilon$ is the region for which the loss function changes from
from being approximately quadratic to being approximately linear.


The total loss of the flow generator is composed of three main parts: the standard GAN loss (for both discriminators), the optical flow reconstruction loss which includes the two weighted losses (EPE, Charbonnier) and the motion warping loss. The resulting loss can be formulated as follows:
\begin{equation}
    \mathcal{L}_{G_{F}} = \lambda_1\mathcal{L}_{GAN}
    + \lambda_2 \mathcal{L}_{epe} + \lambda_3 \mathcal{L}_{char} + \mathbf{1}_{c}\lambda_4\mathcal{L}_{G_{W}}
\end{equation}
where $\lambda_1$, $\lambda_2$, $\lambda_3$ and $\lambda_4$ are scaling factors, $\mathcal{L}_{GAN}=1/2(\mathcal{L}_{GAN}(G_{F}, D_{P})+\mathcal{L}_{GAN}(G_{F}, D_{E}))$ and $\mathbf{1}_{c}$ is the indicator function to indicate whether to connect the two losses, \ie $\mathbf{1}_{c}=0$ if $epoch < e_{c}$ and $1$ otherwise. Indeed, the two generators are trained separately in parallel until a given epoch $e_{c}$ where the image generator comes into play to judge whether the flows frontalized by the flow generator are warpable into the texture. $e_{c}$ indicates a level at which the image generator manages to generate realistic expressive faces.

Regarding the discriminators, we use standard loss functions, namely sigmoid cross-entropy for the first discriminator and categorical cross-entropy for the emotion discriminator.

\section{\ba{Experimental Protocol}}
\label{sec:exp}

\subsection{Datasets}
\label{sub:data}
We conducted our experiments on five datasets covering a large variability of intra-individual facial expressions.
Training our end-to-end model requires a dataset with paired images. We selected SNaP-2DFe \cite{allaert2018impact} as it's the only publicly available dataset with paired images sequences with expressive faces acquired simultaneously with and without head pose variation.
The training phase is assisted with commonly used datasets for FER  including CK+ \cite{lucey2010extended}, ADFES \cite{van2011moving}, MMI \cite{pantic2005web}, Oulu-CASIA \cite{zhao2011facial} datasets. 
These datasets are being included mostly to strengthen the facial motion patterns. 
Regarding those without frontal versus non-frontal views, we created non-frontal images by applying basic rotations and adding random noise to the original images. We used the original images as the expected frontalized ground-truth. The datasets selected ensure a consistent validation protocol. \ba{Other commonly used datasets, such as BP4D \cite{zhang2014bp4d}, are excluded from our training due to the manner in which head pose variations are artificially generated, i.e. rotating the camera around the face in 3D space, bypasses the bio-mechanical constraints of the face, which runs counter to model-based learning.}

Simultaneous Natural and Posed 2D Facial expressions dataset (SNaP-2DFe) \cite{allaert2018impact} is an image dataset that meets the requirements requested for the motion frontalization phase. It contains paired images of frontal and non-frontal faces for different subjects and facial expressions. More specifically, it provides images of subjects with expressive faces acquired simultaneously with head pose variation (unconstrained recording) and without head pose variation (constrained recording) from 15 different subjects with 6 categories of head poses (yaw, pitch, roll, diagonal, nothing and Tx) and 7 different facial expressions most commonly used in the literature (disgust, happiness, anger, surprise, neutral, fear and sadness).



\subsection{Evaluation Protocols}
The evaluation is performed by comparing the results obtained by a facial expression classifier in both motion domain and image domain when no frontalization is applied (original flows and non-frontal images) and when frontalization is applied (frontalized flows and warped frontal images). We emphasize that our end-to-end model improves the FER task in the presence of head movements in both motion and image domains, as it assists the classification models in increasing the recognition accuracy in presence of head pose variation.

The evaluation of our model is performed following a 10-Fold cross-validation (9-fold for training, 1-fold for testing). 
The 9 training folds from different datasets are merged to reconstruct the training samples, and the evaluation is done on the remaining fold of each dataset separately.
Performance is evaluated on the basis of the classification accuracy using different FER models.

\begin{figure}[t]
    \centering
    \includegraphics[width=1.\columnwidth]{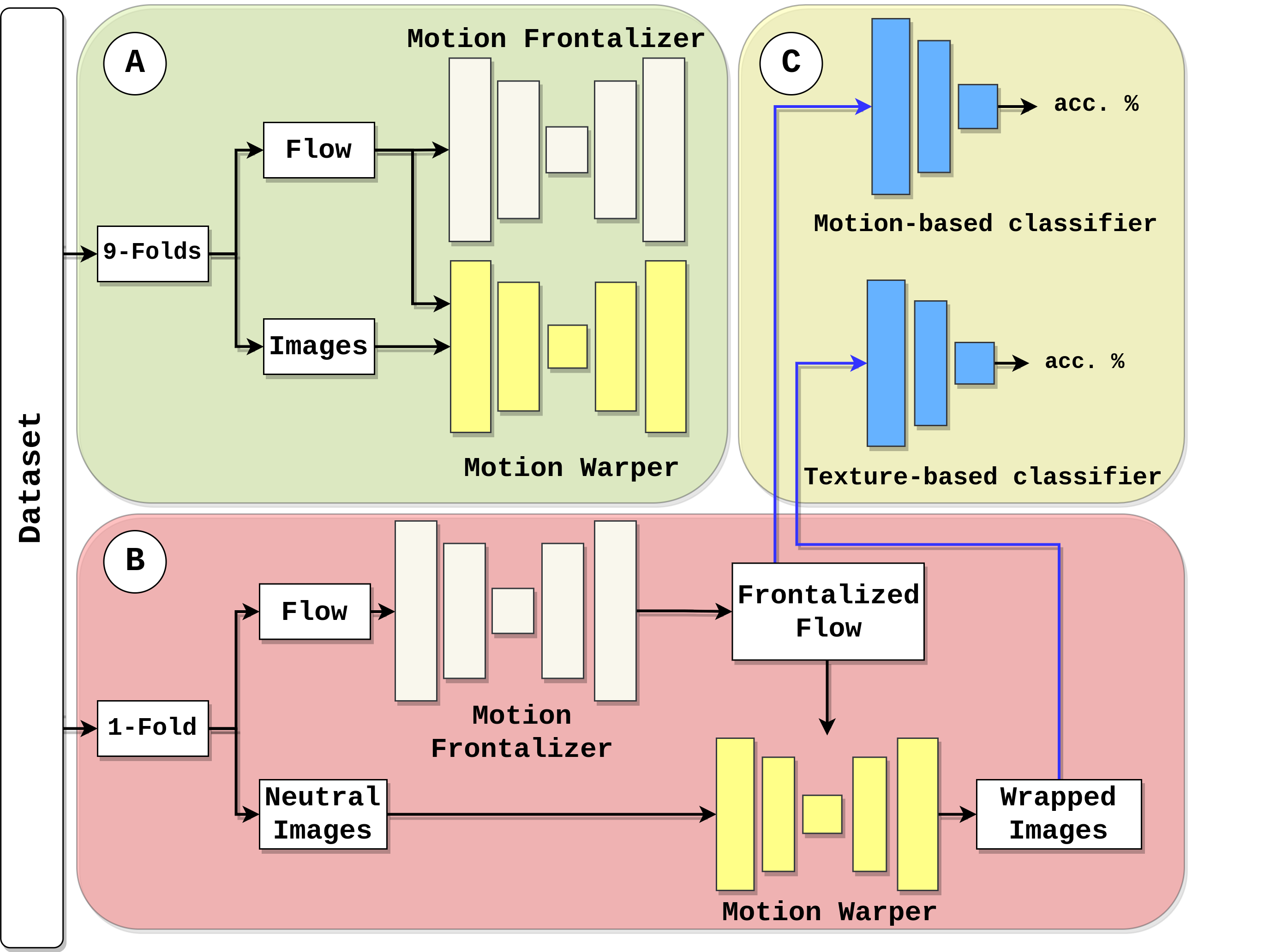}
    \caption{Cross-validation evaluation protocol. A: training the end-to-end model on training folds. B: frontalizing flows and warping images of the test fold using the trained models. C: evaluating performances using FER models on the reconstructed test fold in both motion and image domains.}
    \label{fig:eval_proto}
\end{figure}

The evaluation is performed in three main steps as illustrated in Figure \ref{fig:eval_proto}. First, the frontalization and warping models are trained on the training folds. Then, the trained frontalization model is used to normalize the non-frontal flows in the test folds, and these frontalized flows are used with the corresponding frontal neutral faces to reconstruct the expressive faces with the trained warping model. Finally, the reconstructed flows and images are processed by FER models in both motion and image domains to obtain recognition accuracies. 

In order to provide a large variety of optical flows and to cover a wide range of facial motion intensities, we propose to combine two different ways of calculating optical flows in an image sequence; (1) The \emph{neutral-apex strategy}, it consists in calculating the optical flow between the first frame (neutral) and the last frame (apex) in each image sequence $s$ of length $n$, \ie the set $\{(f_{s_0}, f_{s_n}); \forall s\}$. The larger $n$ is, the larger the head pose and thus the higher the intensity of the motion, and conversely. (2) The \emph{frame-to-frame strategy}, which consists of using all the frames in each sequence $s$ by calculating the optical flow between each pair of consecutive frames, \ie the set $\{(f_{s_i}, f_{s_{i+1}}); \forall i, s\}$. The motion intensity depends on the time interval between the frames. Yet, for this strategy, the motion is less intense than for the neutral-apex strategy. 

\subsection{Implementation Details}

The facial motion is calculated using Farneback method \cite{farneback2003two} which extracts optical flow from grayscale images. Allaert et al. have shown it to be best suited to FER \cite{allaert2022comparative}.
We use a clipping value of $10.0$ for flow normalization for magnitude calculated according to the average distribution of the facial movement and motion intensities on SNaP-2DFe dataset, and we normalize both magnitudes and directions (in radian).

For the motion frontalization model, important learning parameters include the learning rates. We selected Adam with $1\times10^{-5}$ for the flow generator, Adam with $1\times10^{-4}$ for the patches discriminator and SGD with $1\times10^{-4}$ for the expression discriminator. The learning rates are gradually decreased during training.
 We follow DCGAN \cite{radford2015unsupervised} and use a slope (alpha parameter) of 0.2 for the patches discriminator.
The loss function scaling factors, are set as follows: $\lambda_1=1$, $\lambda_2=7$, $\lambda_3=3$ and $\lambda_4=1$. 

For the motion warping model, we used Adam and set the learning rate to $10^{-5}$ for the image generator, the learning is reduced accordingly as well.
The loss function scaling factors are set as follows: $\alpha_1=2$, $\alpha_2=7$ and $\alpha_3=1$.

Our model is trained for 15 epochs with a batch size of 4. The connection epoch between the two losses of the flow generator and the image generator is set to $e_c=5$. We use the PyTorch libraries \cite{paszke2017automatic} to implement our end-to-end model.  


\section{\ba{Evaluations}}
\label{sec:eval}
We assess the effectiveness of our method through two sets of experiments. First, we compare our approach with recent FVS methods. Secondly, we illustrate its potential for cross-subject facial motion transfer. Finally, we conduct an ablation study to investigate the contribution of each loss term to the overall performance of the model.

\subsection{Frontal View Synthesis}
\label{sub:FVS}
We propose a qualitative and a quantitative evaluation in both motion and image domains. First, we demonstrate the efficiency of our end-to-end model to generate frontal faces while preserving facial expressions.
Then we compare our approach against different FVS approaches and show its efficiency in reducing the FER gap between frontal and non-frontal faces when using different FER methods. 

We compare our approach against the recent best FVS approaches. For hand-crafted approaches, we include SIFT flow \cite{liu2008sift} (non-rigid registration based on optical flow), while for deep learning approaches, we consider CFR-GAN \cite{ju2022complete} and pixel2style2pixel (pSp) \cite{richardson2021encoding} for image-based methods, and GLU-Net \cite{truong2020glu} 
for optical-flow-based methods. 

In order to make a fair comparison between the different FVS approaches, all experiments were performed under identical settings,
thus ensuring that any biases that might result from model selection or classifier optimization are omitted. The baseline involves only face detection, followed by cropping around the face, that is, no frontalization is applied, only the most basic pre-processing step. All approaches are applied to the baseline images. We compare the results in both motion and image domains.

\subsubsection{Qualitative Comparison}
We first conduct a qualitative experiment on evaluation datasets. We compare the FVS results of our proposed method for both motion and texture.

For our method, the flow frontalization is done directly in the motion domain, but for those for which the frontalization is done on the image, we first frontalize the image and then compute the motion on the frontalized faces. The comparison is made on the basis of three common metrics, structural similarity (SSIM) and RMSE for images, and EPE for optical flows. Table \ref{tab:metrics} shows the superiority of our method in both motion and image over the other FVS methods. 
Indeed, a lower EPE indicates that our frontalization model succeeds in frontalizing motion while preserving facial expression patterns, which is mainly due to the fact that our method frontalizes motion rather than image, allowing it to take advantage of the similarity of expression patterns between different subjects, and a higher SSIM indicates that the warping model successfully warps the frontalized motion in the image, which is mainly driven by the perceptual loss included in the warping model that allows it to generate consistent expressive faces while preserving the facial identity of the subject.

\begin{figure}[t]
    \centering
    \includegraphics[width=1.\columnwidth]{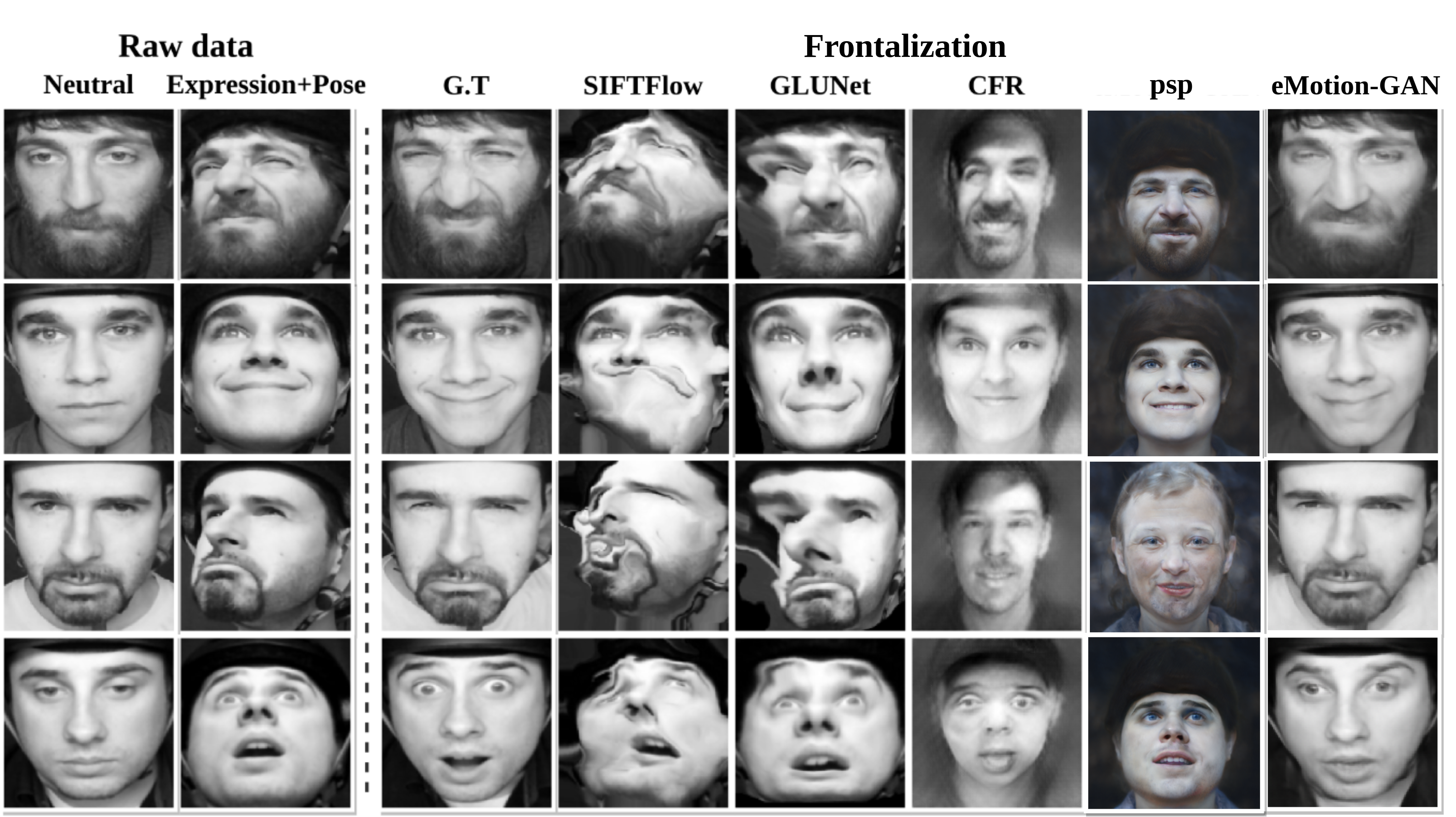}
    \caption{Qualitative comparison of different FVS methods in image domain. Our approach effectively handles large variations for frontal view synthesis while preserving facial expressions.}
    \label{fig:qual_comparison_texture}
\end{figure}

For image warping, Figure \ref{fig:qual_comparison_texture} shows the results given by the different FVS methods on expressive faces with head pose variations. In terms of image reconstruction, our method manages to warp the frontalized flow into the neutral face to recover the original facial expression. Unlike other methods, our method does not suffer from artifacts and distortion of expression patterns. Thus, allowing the facial expression to be recovered without loss of identity of the subject.

\subsubsection{FER Comparison}
\label{sub:fer_comparison}
We perform FER comparison between the FVS methods cited in Section \ref{sec:eval} in both motion and image.
We denote by $\Delta$(\%) the percent improvement in recognition accuracy between non-frontalized (original non-frontal faces) and frontalized faces (synthesized faces).

\begin{table}[!t]
    \centering
    \caption{Mean and standard deviation of EPE, SSIM and RMSE across validation datasets. $\downarrow$ indicates that the lower the result, the better it is and vice versa, the best results are shown in bold.}\label{tab:metrics}
    \fontsize{9.5}{9.5}\selectfont
        \begin{tabular}{lcccccc}
            \toprule
             Method   & EPE $\downarrow$ & SSIM $\uparrow$ & RMSE $\downarrow$\\
             \midrule
             SIFT \cite{liu2008sift}  &  36.07$\pm$13.50         & 0.58$\pm$0.13          & 9.10$\pm$0.70           \\
             CFR \cite{ju2022complete}   &  44.96$\pm$6.24          & 0.31$\pm$0.07          & 9.99$\pm$0.11           \\
             GLUNet \cite{truong2020glu} &  33.93$\pm$6.58          & 0.66$\pm$0.13          & \textbf{7.62}$\pm$ 1.04 \\
             pSp \cite{richardson2021encoding} &  40.32$\pm$6.01    & 0.36$\pm$0.07         & 9.67$\pm$0.10 \\
             Ours   &  \textbf{14.34}$\pm$3.99 & \textbf{0.78}$\pm$0.18 & 7.75$\pm$1.00           \\
            \bottomrule 
        \end{tabular}
\end{table}

\begin{table*}
    \centering
    \caption{Quantitative comparison of different FVS methods using different FER methods in both image (top part) and optical flow (bottom part) on validation datasets. The best image scores are underlined and the best optical flow scores are in bold for each dataset. For each approach, the 5 rows correspond to the scores obtained with the use of the different FVS methods in the following order: Baseline (no frontalization), SIFT flow\cite{liu2008sift}, CFR-GAN \cite{ju2022complete}, GLUNet \cite{truong2020glu}, pSp \cite{richardson2021encoding} and our eMotion-GAN.}\label{tab:FER_results_comparison}
    \fontsize{8.5}{9.8}\selectfont
    \begin{tabular}{|c|cccccccccccccc}
        \toprule
        \multicolumn{1}{c}{} & \multicolumn{1}{c}{FER} & \multicolumn{2}{c}{CK+} & \multicolumn{2}{c}{ADFES} & \multicolumn{2}{c}{MMI} & \multicolumn{2}{c}{CASIA} & \multicolumn{2}{c}{SNAP-F}  &  \multicolumn{2}{c}{SNAP-NF}   \\
        \multicolumn{1}{c}{} & Method & acc.[\%] & std[\%] & acc.[\%] & std[\%] & acc.[\%] & std[\%] & acc.[\%] & std[\%] & acc.[\%] & std[\%] & acc.[\%] & std[\%]\\
        \midrule
        \midrule
         \parbox[t]{2mm}{\multirow{20}{*}{\rotatebox[origin=c]{90}{Image-based}}} & 
         \multirow{5}*{Ad-Corre \cite{9727163}} 
         & 80.38 & $\pm$5.38 & 90.93 & $\pm$8.02 & 70.07  & $\pm$11.79 & 62.77 & $\pm$7.42 & 59.19 & $\pm$2.24 & 58.13 & $\pm$5.08\\
         & & 78.66 & $\pm$7.00 & 89.29 & $\pm$9.87 & 68.05  & $\pm$10.47 & 61.31 & $\pm$7.30 & 55.77 & $\pm$4.48 & 47.03 & $\pm$5.24\\
         & & 68.34 & $\pm$4.86 & 76.32 & $\pm$7.29 & 61.71  & $\pm$11.84 & 54.80 & $\pm$3.52 & 37.83 & $\pm$3.35 & 45.09 & $\pm$3.98\\
         & & 79.80 & $\pm$6.53 & 86.32 & $\pm$8.81 & 69.05  & $\pm$12.24 & 60.88 & $\pm$6.25 & 53.63 & $\pm$4.52 & 47.65 & $\pm$5.34\\
         & & 65.83 & $\pm$4.4 & 69.4 & $\pm$7.08 &  46.36 & $\pm$8.17 & 34.11 & $\pm$4.34 & 47.64 & $\pm$7.35 & 41.66 & $\pm$3.85\\
         & & 77.02 & $\pm$8.53 & 83.96 & $\pm$10.59 & 64.71 &$\pm$ 9.20  & 60.27 & $\pm$4.79 & 60.25 & $\pm$4.92 & 46.35 & $\pm$6.16\\
         \cline{2-14}
         & \multirow{5}*{DAN \cite{biomimetics8020199}}
         & 81.50 & $\pm$5.08 & 86.32 & $\pm$5.51 & 70.12 & $\pm$11.48 & 63.20 & $\pm$8.88 & \underline{71.17} & $\pm$4.41 & \underline{67.30} & $\pm$6.75\\
         & & 74.48 & $\pm$7.75 & 87.03 & $\pm$4.88 & 70.55 & $\pm$8.83  & 62.17 & $\pm$9.51 & 69.46 & $\pm$4.87 & 56.42 & $\pm$4.86\\
         & & 71.43 & $\pm$5.88 & 83.13 & $\pm$6.78 & 62.26 & $\pm$10.79 & 60.27 & $\pm$6.31 & 40.62 & $\pm$5.75 & 49.59 & $\pm$5.17\\
         & & 77.55 & $\pm$6.53 & 83.24 & $\pm$4.42 & 66.19 & $\pm$7.92  & 61.53 & $\pm$9.12 & 68.82 & $\pm$4.80 & 60.06 & $\pm$7.84\\
         & & 69.44 & $\pm$5.1 & 71.81 & $\pm$6.57 & 52.14  & $\pm$7.43 & 44.97 & $\pm$4.48 & 54.06 & $\pm$3.81 & 51.9 & $\pm$4.32\\
         & & 72.55 & $\pm$4.82 & 84.01 & $\pm$7.90 & 65.21 & $\pm$5.14  & 61.33 & $\pm$6.41 & 60.04 & $\pm$5.80 & 45.07 & $\pm$5.47\\
         \cline{2-14}
         & \multirow{5}*{HSE \cite{savchenko2022video}}
         & 79.25 & $\pm$4.94 & \underline{93.90} & $\pm$5.74  & 71.98 & $\pm$8.36 & 63.00 & $\pm$7.58 & 67.53 & $\pm$3.22 & 63.90 & $\pm$6.84 \\
         & & 74.77 & $\pm$4.84 & 90.88 & $\pm$8.90  & 69.57 & $\pm$6.01 & 59.85 & $\pm$7.48 & 62.84 & $\pm$3.39 & 51.72 & $\pm$4.45\\
         & & 69.76 & $\pm$3.77 & 82.47 & $\pm$6.81  & 59.31 & $\pm$9.93 & 56.92 & $\pm$3.58 & 39.53 & $\pm$3.93 & 43.16 & $\pm$4.64\\
         & & 76.74 & $\pm$4.72 & 90.16 & $\pm$8.91  & 65.64 & $\pm$6.49 & 60.27 & $\pm$8.41 & 65.17 & $\pm$3.34 & 58.13 & $\pm$5.87\\
         & & 68.63 & $\pm$4.72 & 83.24 & $\pm$9.51 & 59.4 & $\pm$6.68 & 49.38 & $\pm$7.58 & 52.96 & $\pm$5.38 & 41.23 & $\pm$5.8\\
         & & 75.63 & $\pm$5.91 & 87.14 & $\pm$11.11 & 69.02 & $\pm$7.18 & 61.52 & $\pm$6.25 & 59.20 & $\pm$6.29 & 42.50 & $\pm$4.35\\
         \cline{2-14}
         & \multirow{5}*{DMUE \cite{she2021dive}}
         & \underline{85.71} & $\pm$4.10 & 92.31 & $\pm$4.87 & \underline{74.36} & $\pm$6.96 & \underline{70.70} & $\pm$6.42 & 69.66 & $\pm$1.77 & 66.90 & $\pm$5.23\\
         & & 81.50 & $\pm$4.84 & 88.52 & $\pm$5.20 & 73.40 & $\pm$7.86 & 64.66 & $\pm$7.96 & 68.16 & $\pm$4.12 & 57.07 & $\pm$4.05\\
         & & 77.57 & $\pm$4.78 & 82.42 & $\pm$4.99 & 67.07 & $\pm$7.73 & 66.31 & $\pm$5.38 & 42.10 & $\pm$4.00 & 50.86 & $\pm$5.36\\
         & & 84.59 & $\pm$4.58 & 91.59 & $\pm$4.16 & 72.40 & $\pm$7.94 & 67.36 & $\pm$7.20 & 68.59 & $\pm$2.06 & 61.58 & $\pm$7.94\\
         & & 76.71 & $\pm$6.61 & 86.26 & $\pm$8.27 & 62.79  & $\pm$5.87 & 56.06 & $\pm$5.65 & 60.46 & $\pm$5.15 & 43.59 & $\pm$7.0\\
         & & 83.45 & $\pm$3.06 & 88.63 & $\pm$8.39 & 68.07 & $\pm$8.10 & 70.51 & $\pm$4.16 & 67.53 & $\pm$3.32 & 50.85 & $\pm$4.50\\
        \midrule
        \addlinespace[1.0ex]
        \midrule
        \parbox[t]{2mm}{\multirow{5}{*}{\rotatebox[origin=c]{90}{Flow-based}}} & 
         \multirow{5}*{Flow-CNN}
         & 86.27 & $\pm$6.08 & 90.83 & $\pm$7.96  & 69.48 & $\pm$10.83 & 69.89 & $\pm$5.95 & \textbf{86.55} & $\pm$3.59 & 42.31 & $\pm$5.76\\
         & & 75.88 & $\pm$4.77 & 76.21 & $\pm$11.81 & 57.40 & $\pm$10.87 & 59.80 & $\pm$6.74 & 67.28 & $\pm$7.81 & 33.96 & $\pm$5.44\\
         & & 69.73 & $\pm$6.58 & 76.26 & $\pm$8.59  & 53.02 & $\pm$12.74 & 56.03 & $\pm$6.38 & 43.12 & $\pm$7.01 & 35.66 & $\pm$5.68\\
         & & 80.63 & $\pm$4.47 & 86.32 & $\pm$7.75  & 67.02 & $\pm$11.89 & 66.52 & $\pm$7.18 & 81.19 & $\pm$4.78 & 54.91 & $\pm$3.35 \\
         & & 70.15 & $\pm$6.13 & 74.17 & $\pm$8.41 & 52.06 & $\pm$12.96 & 56.03 & $\pm$6.21 & 60.01 & $\pm$6.39 & 45.84 & $\pm$5.43\\
         & & \textbf{90.74} & $\pm$3.01 & \textbf{94.62} & $\pm$5.19 & \textbf{75.79} & $\pm$9.81 & \textbf{73.22} & $\pm$5.64 & 80.55 & $\pm$6.21 & \textbf{64.10} & $\pm$5.74\\
         \midrule
        \bottomrule
    \end{tabular}
\end{table*}

For motion-based FER, we compare the recognition accuracy given by the motion-based CNN (Flow-CNN) detailed in Section \ref{sec:approach} on different evaluation datasets. To our knowledge, no pre-trained model has been proposed for FER in optical flow, therefore we use our proposed motion-based CNN (expression discriminator) to classify the frontalized flows.

For FER on warped images, we cite the latest research advances in the field of FER to evaluate the expressive faces given by different FVS approaches.
We include DMUE \cite{she2021dive}, a FER method that relies on latent distribution mining to estimate the pairwise uncertainty between well-labeled and ambiguous samples.
HSEmotion \cite{savchenko2022video} extracts facial features with the single EfficientNet \cite{tan2019efficientnet} model pre-trained on AffectNet \cite{mollahosseini2017affectnet}.
DAN \cite{biomimetics8020199} is an attention-based approach for FER.
We also cite Ad-Corre \cite{9727163}, an approach that employs adaptive correlation loss to maximize correlation for intra-class samples and minimize it for inter-class samples.
We use the pre-trained models provided by the authors and we emphasize that all these methods are designed to deal with FER in the wild. DMUE, DAN and Ad-Corre are trained on AffectNet dataset \cite{mollahosseini2017affectnet} and HSEmotion is trained on AFEW dataset \cite{kossaifi2017afew}.

Table \ref{tab:FER_results_comparison} presents the results obtained by different FVS methods when using different FER methods in both motion and image. 
In image domain, the baseline seems to achieve the best results 
given that recent FER approaches are trained to analyze frontal and less constrained faces, thanks to augmentation methods. Image-based frontalization tends to introduce excessive deformation, altering both facial structure and expression. Our motion-based approach applies a loss that preserves face structure and ensures consistent textural rendering, albeit with a slight attenuation of expression (see Figure \ref{fig:qual_comparison_texture}).
Nevertheless, our method is still competitive on most datasets as it nearly achieves the results given by recent FER methods in the literature.
However, in motion field, the results clearly demonstrate that our proposed method outperforms other FVS methods across datasets. 
The percent improvement $\Delta$ in optical flow attains +5\% on small pose variations and up to +20\% on large pose variations.
While our method is invariant in frontal settings, \eg CK+ and ADFES.

These results highlight the strength of motion-based approaches over image-based approaches, where inter-individual variations tend to impact the quality of frontalization and thus reduce FER performance. Furthermore, unlike image-based methods that hardly manage to handle both frontal and non-frontal settings, motion-based methods achieve high performance in non-frontal settings while being invariant in frontal settings.

\subsection{Cross-subject Facial Motion Transfer}
\begin{figure}[t]
    \centering
    \includegraphics[width=1.\columnwidth]{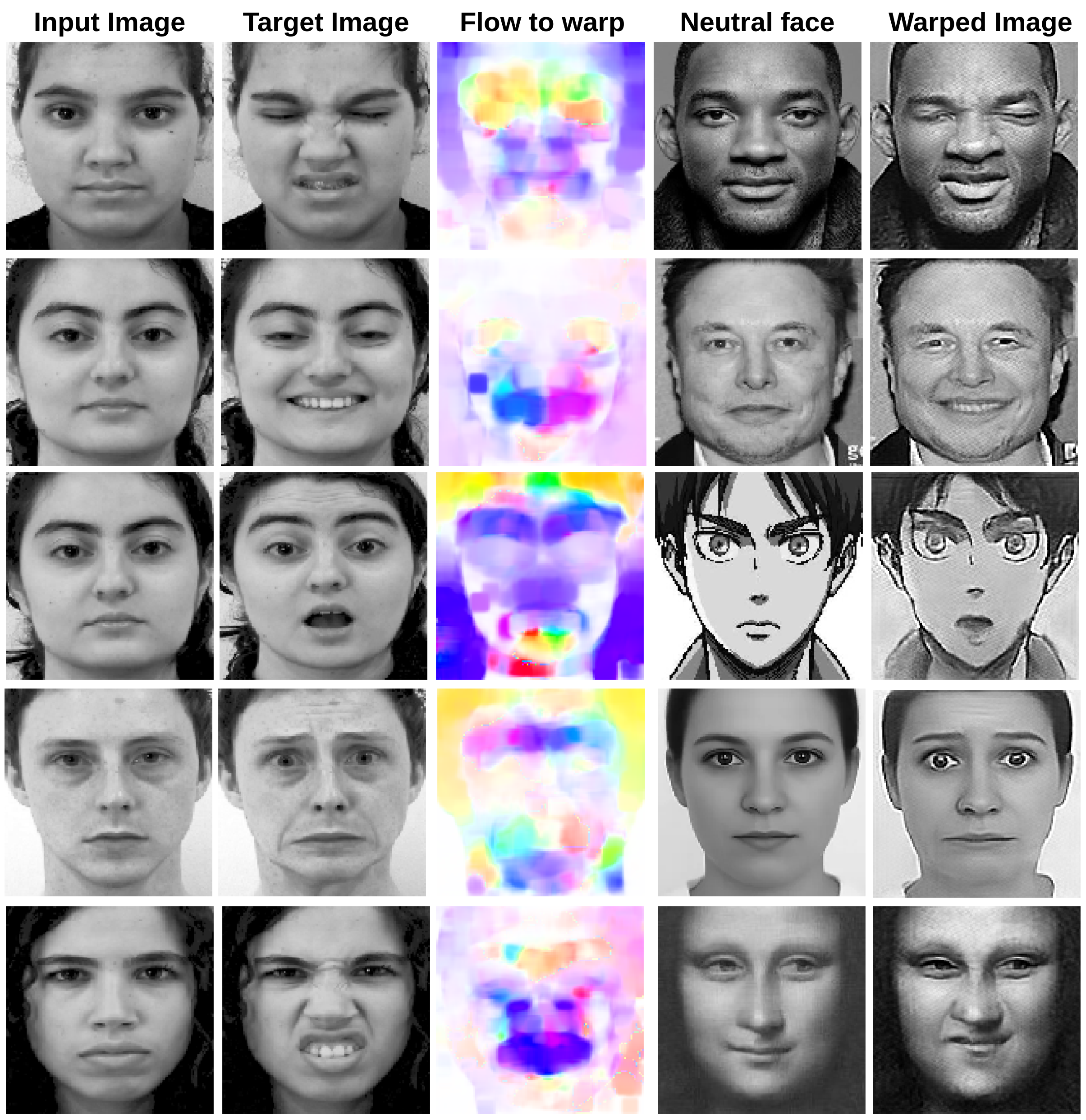}
    \caption{Cross-subject motion transfer: Our motion warping model transfers motion patterns between faces while preserving identity.}
    \label{fig:motion_transfer}
\end{figure}

In this section, we carry out an experiment to demonstrate the effectiveness of our motion warping model for transferring facial motion between two different subjects. In other words, our model can transfer facial expressions from subject A to subject B by warping a motion field, corresponding to the facial expression, from subject A to subject B's face.
We emphasize that the preliminary face registration step guarantees consistency in facial morphology.
Figure \ref{fig:motion_transfer} shows the results of cross-subject motion transfer between different individuals.

Our motion warping model can be used for different interesting applications, including: 1) data augmentation by generating a variety of expressive faces given a neutral face. Indeed, along with our warping model, we can build a conditioned GAN to generate facial expressions motion conditioning on the facial expression and warp the generated facial motion to neutral faces to generate additional data. 2) class balancing: to balance number of samples per expression in a non-balanced FER dataset. 3) intensification or attenuation of facial expressions: to transform micro-expressions into macro-expressions and vice versa. 4) category balancing: as in many datasets, categories such as gender, age and ethnicity are not balanced, our warping model can be used to overcome such constraint.

We emphasize that our warping model can be used to warp a facial motion field into any frontal face, whether real or not.
For instance, in real-world applications, \eg surveillance cameras, it is challenging to obtain the neutral frontal face of a subject. However, the frontalized motion can be warped into any neutral frontal face to get the corresponding expression.

For a quantitative evaluation, we propose to warp facial expressions motion from different datasets into a neutral average face (raw 4, column 4 in Figure \ref{fig:motion_transfer}) and then calculate the FER accuracies on the warped faces. 
The FER results are provided in Table \ref{tab:motion_transfer}. 

\begin{table}[!b]
    \centering
    \caption{FER accuracies on warped images across evaluation datasets using average face.}\label{tab:motion_transfer}
    \fontsize{9}{10}\selectfont
        \begin{tabular}{lccccc}
            \toprule
            \textbf{Method} &  CK+ & ADFES & MMI & CASIA & SNAP-F\\
            \midrule
            Ad-Corre  & 75.33\% & 85.49\% & 59.90\% & 52.92\% & 55.33\%\\
            DAN       & 71.17\% & \textbf{93.90\%} & 60.31\% & 56.68\% & 59.85\%\\
            HSE       & 68.33\% & 89.40\% & 57.50\% & 54.61\% & 54.09\% \\
            DMUE      & \textbf{77.31\%} & 88.52\% & \textbf{63.74\%} & 60.03\% & \textbf{61.12\%} \\
            \bottomrule
        \end{tabular}
\end{table}

These results highlight the ability of our motion warping model to cross-subject facial expression transfer.
As a matter of fact, it manages to obtain scores close to those obtained on the original faces (\eg $\sim$94\% on ADFES).

\subsection{Ablation Study}

In this section, we conduct an ablation study on eMotion-GAN, analyzing variants with one main loss discarded at a time (normalization model loss, emotion discriminator loss, and warping model loss). This allows us to understand the distinct contributions of each loss during training. We use the same end-to-end model, parameters, and training protocol, presenting accuracy results for various scenarios (motion and image) on different datasets in Table \ref{tab:ablation_study}.

\paragraph{without motion reconstruction loss ($\mathcal{L}_{epe}$ and $\mathcal{L}_{char}$)}
The absence of the two reconstruction losses, $\mathcal{L}_{epe}$ and $\mathcal{L}_{char}$, during training leads to a deficiency in the key details of facial expression in both motion and texture. Consequently, classifiers struggle to effectively distinguish between different expressions, resulting in less satisfactory outcomes in both normalized flows and synthesized faces. It is worth pointing out that reconstruction loss is of significant importance as a central component of our model.

\paragraph{without emotion discriminator loss ($\mathcal{L}_{D_E}$)}
When the loss of the emotion discriminator is not included in model training, the model lacks crucial feedback on the main signal that needs to be preserved, i.e. the motion of the expression. Consequently, this absence has a significant impact on the warping process, leading to sub-optimal results and compromising overall model performance.

\paragraph{without warping model loss ($\mathcal{L}_{G_W}$)}
When the warping model loss is retained as part of the total loss for the motion normalization model, motion flows of the generated frontalized faces are produced; however, they do not exhibit optimal warpability when applied to neutral faces. In other words, the warping process does not effectively align the motion flows to the neutral face, leading to poor results in the overall frontalization process.

\begin{table}[!t]
    \centering
    \caption{Ablation study on different losses with FER accuracies in both image (top line) and optical flow (bottom line). Best image scores are underlined, best flow scores are in bold.}\label{tab:ablation_study}
    \fontsize{9}{10}\selectfont
        \begin{tabular}{lcccc}
            \toprule
            \textbf{Method} & ADFES & MMI & CASIA & SNAP-NF\\
            \midrule
            \multirow{2}*{Ours w/o $\mathcal{L}_{epe},\mathcal{L}_{char}$} & 30.14\% & 38.10\% & 38.75\% & 28.15\% \\
             & 44.12\% & 40.05\%  & 42.08\% & 30.29\%\\
             \hline
            \multirow{2}*{Ours w/o {$\mathcal{L}_{D_E}$}} & 71.43\% & 47.62\% & 45.58\% &  38.30\%\\
             & 92.86\% & 66.67\% & 68.75\% & 51.06\%\\
             \hline
            \multirow{2}*{Ours w/o $\mathcal{L}_{G_W}$} & 73.14\% & 45.95\% & 55.83\% & 40.43\% \\
             & 92.86\% & 70.95\% & 70.83\% & 61.70\% \\
             \hline
            \multirow{2}*{Ours} & \underline{88.63\%} & \underline{68.07\%} & \underline{70.51\%} & \underline{50.85\%}\\
             & \textbf{94.62\%} & \textbf{75.79\%} & \textbf{73.22\%} & \textbf{64.10\%} \\
            \bottomrule
        \end{tabular}
\end{table}

\section{Conclusions}
\label{sec:conclusion}
We propose eMotion-GAN, a deep learning approach for frontal view synthesis that jointly relies on facial motion analysis and cGANs to correct head pose variations for FER. Our approach involves a motion frontalization model and a motion warping model, which successfully reconstructs expressive frontal faces from facial motion patterns.
Extensive experiments demonstrate the ability of our model to frontalize the face while preserving facial expressions.
Unlike image-based methods that hardly manage to handle both frontal and non-frontal settings, our motion-based method achieve high performance in non-frontal settings while being invariant in frontal settings.
The FER percent improvement in motion reaches +5\% on small poses and up to +20\% on large poses.




Finally, while our primary goal is to generate frontal views while preserving facial expressions, it's important to recognize the potential misuse of our approach in deepfake related contexts.

\bibliographystyle{IEEEtran}
\bibliography{bib}

\vfill

\end{document}